\RequirePackage{fix-cm}
\documentclass[smallextended]{svjour3}       
\smartqed  
\usepackage{graphicx}
\usepackage{amsmath, url}
\usepackage[english, francais]{babel}
\usepackage[utf8x]{inputenc}
\usepackage{graphicx}
\usepackage{algpseudocode} 
\usepackage{algorithm}
\usepackage{algorithmicx}
\usepackage{multirow}
\usepackage[table,xcdraw]{xcolor}

\begin{document}

\title{Automatic Quality Assessment for Speech Translation Using Joint ASR and MT Features}


\titlerunning{Automatic Quality Assessment for Speech Translation}

\author{Ngoc-Tien Le \and Benjamin Lecouteux  \and Laurent Besacier
}

%

\authorrunning{Ngoc-Tien Le et al.}

\institute{Firstname Lastname 
	\at Laboratoire d'Informatique de Grenoble, University of Grenoble Alpes, France\\
	Building IMAG, 700 Centrale, 38401 Saint Martin d'Hères \\
	Tel.: +33 457421454 \\
	\email{firstname.lastname@imag.fr}
}

\date{Received: date / Accepted: date}

\maketitle

\selectlanguage {english}
\begin{abstract}

This paper addresses automatic quality assessment of spoken language translation (SLT). This relatively new task is defined and formalized as a sequence labeling problem where each word in the SLT hypothesis is tagged as $good$ or $bad$ according to a large feature set. We propose several word confidence estimators (WCE) based on our automatic evaluation of transcription (ASR) quality, translation (MT) quality, or both (combined ASR+MT). This research work is possible because we built a specific corpus which contains 6.7k utterances for which a quintuplet containing: ASR output, verbatim transcript, text translation, speech translation and post-edition of translation is built.
The conclusion of our multiple experiments using joint ASR and MT features for WCE is that MT features remain the most influent while ASR feature can bring interesting complementary information. 
Our robust quality estimators for SLT can be used for re-scoring speech translation graphs or for providing feedback to the user in interactive speech translation or computer-assisted speech-to-text  scenarios.

 \and 
\end{abstract}

\section{Introduction}
\label{sec:intro}

Automatic quality assessment of spoken language translation (SLT), also named confidence estimation (CE), is an important topic because it allows to know if a system produces (or not) user-acceptable outputs. In interactive speech to speech translation, CE helps to judge if a translated turn is uncertain (and ask the speaker to rephrase or  repeat). For speech-to-text applications, CE may tell us if output translations are worth being corrected or if they require retranslation from scratch. 
Moreover, an accurate CE can also help to improve SLT itself through a second-pass N-best list re-ranking or search graph re-decoding, as it has already been done for text translation in \cite{bach11} and \cite{luong:hal-00953719}, or for speech translation in \cite{besacier-asru-2015}.
Consequently, building a method which is capable of pointing out the correct parts as well as detecting the errors in a speech translated output is crucial to tackle above issues.


Given signal $x_{f}$ in the source language, spoken language translation (SLT) consists in finding the most probable target language sequence $\hat{e} = (e_{1}, e_{2},...,e_{N})$ so that

\begin{equation}
\hat{e}=\underset{e}{\operatorname{argmax} } \{p(e/x_{f},f)\}
\end{equation}
where $f = (f_{1}, f_{2},...,f_{M})$ is the transcription of $x_{f}$.

Now, if we perform confidence estimation at the ``words'' level, the problem is called Word-level Confidence Estimation (WCE) and we can represent this information as a sequence $q$ (same length $N$ of $\hat{e}$) where $q = (q_{1}, q_{2},...,q_{N})$ and $q_{i}\in\{good,bad\}$\footnote{$q_{i}$ could be also more than 2 labels, or even scores but this paper only
deals with error detection (binary set of labels)}.

Then, integrating automatic quality assessment in our SLT process can be done as following:

%
%

\begin{eqnarray}
\hat{e} &=& \underset{e}{\operatorname{argmax} } \sum_{q}{p(e,q/x_{f},f)} \\
\hat{e} &=& \underset{e}{\operatorname{argmax} } \sum_{q}{p(q/x_{f},f,e)*p(e/x_{f},f)} \\
\hat{e} &\approx& \underset{e}{\operatorname{argmax} }  \{ \underset{q}{\operatorname{max} }  \{ p(q/x_{f},f,e)*p(e/x_{f},f) \} \}
\end{eqnarray}

In the product of (4), the SLT component $p(e/x_{f},f)$ and the WCE component $p(q/x_{f},f,e)$ contribute together to find the best translation output $\hat{e}$.
 In the past, WCE has been treated separately in ASR or MT contexts and we propose here a joint estimation of word confidence for a spoken language translation (SLT) task involving both ASR and MT. 
 
This journal paper is an extended version of a paper published at ASRU 2015 last year \cite{besacier-asru-2015} but we focus more on the WCE component and on the best approaches to estimate $p(q/x_{f},f,e)$ accurately.

\vspace{0.2cm}
\textbf{Contributions} The main contributions of this journal paper are the following:

\begin{itemize}
\item A corpus (distributed to the research community\footnote{https://github.com/besacier/WCE-SLT-LIG}) dedicated to WCE for SLT was initially published in \cite{besacier14}. We present,  in this paper, its extension from 2643 to 6693 speech utterances. 
\item While our previous work on quality assessment was based on two separate WCE classifiers (one for quality assessment in ASR and one  for quality assessment in MT), we propose here a unique \textit{joint} model based on different feature types (ASR and MT features).
\item This \textit{joint} model allows us to operate feature selection and analyze which features (from ASR or MT) are the most efficient for quality assessment in speech translation.
\item We also experiment with two ASR systems that have different performance in order to analyze the behavior of our SLT quality assessment algorithms at different levels of word error rate (WER).
\end{itemize}


\vspace{0.2cm}
\textbf{Outline} The outline of this paper goes simply as follows: section \ref{sec:rw} reviews the state-of-the-art on  confidence estimation for ASR and MT. 
Our WCE system using multiple features is then described in section ~\ref{sec:WCE}. The experimental setup (notably our specific WCE corpus) is presented in section~\ref{sec:db} while section~\ref{sec:WCEjoint}  evaluates our joint WCE system. Feature selection for quality assessment in speech translation is analyzed in section \ref{subsec:featSel} and finally, section \ref{sec:illust}
 concludes this work and gives some perspectives.

\section{Related work on confidence estimation for ASR and MT}
\label{sec:rw}

Several previous works tried to propose effective confidence measures in order to detect errors on ASR outputs. Confidence measures are introduced for Out-Of-Vocabulary (OOV) detection by \cite{Asadi1990}. \cite{Young1994} extends the previous work and introduces the use of word posterior probability (WPP) as a confidence measure for speech recognition. Posterior probability of a word is most of the time computed using the hypothesis word graph \cite{Kemp1997}. Also, more recent approaches \cite{Lecouteux2009} for confidence measure estimation use side-information extracted from the recognizer: normalized likelihoods (WPP), the number of competitors at the end of a word (hypothesis density), decoding process behavior, linguistic features, acoustic features (acoustic stability, duration features) and semantic features.

In parallel, the Workshop on Machine Translation (WMT) introduced in 2013 a WCE task for machine translation. \cite{han13} \cite{luong13b} employed the Conditional Random Fields (CRF) \cite{lafferty01} model as their machine learning method to address the problem as a sequence labelling task. Meanwhile, \cite{bicici13} extended their initial proposition by dynamic training with adaptive weight updates in their  neural network classifier. As far as prediction indicators are concerned, \cite{bicici13} proposed seven word feature types and found among them the ``common cover links'' (the links that point from the leaf node containing this word to other leaf nodes in the same subtree of the syntactic tree) the most outstanding. \cite{han13} focused only on various n-gram combinations of target words. Inheriting most of previously-recognized features, \cite{luong13b} integrated a number of new indicators relying on graph topology, pseudo reference, syntactic behavior (constituent label, distance to the semantic tree root) and polysemy characteristic. The estimation of the confidence score uses mainly 
classifiers like Conditional Random Fields \cite{han13,luong_wmt14}, 
Support Vector Machines \cite{langlois12} or Perceptron \cite{bicici13}.
Some investigations were also conducted to determine which features seem to 
be the most relevant. \cite{langlois12} proposed to filter features using a 
forward-backward algorithm to discard linearly correlated features. Using 
Boosting as learning algorithm, \cite{luong15} was able to take advantage 
of the most significant features.

Finally, several toolkits for WCE were recently proposed: \textit{TranscRater} for ASR\footnote{https://github.com/hlt-mt/TranscRater}, Marmot for MT\footnote{https://github.com/qe-team/marmot} as well as WCE-LIG \cite{servan-toolkit-2015}\footnote{https://github.com/besacier/WCE-LIG} that will be used to extract MT features in the experiments of this journal paper.

To our knowledge, the first attempt to design WCE for speech translation, using both ASR and MT features, is our own work  \cite{besacier14,besacier-asru-2015}   which is further extended in this journal paper submission.

\section{Building an efficient quality assessment (WCE) system}
\label{sec:WCE}
The WCE component solves the equation:

%

\begin{equation}
\label{eq:wce_component}
\hat{q}=\underset{q}{\operatorname{argmax} } \{p_{SLT}(q/x_{f},f,e) \}
\end{equation}

where $q = (q_{1}, q_{2},...,q_{N})$ is the sequence of quality labels on the target language. This is a sequence labelling task that can be solved with several machine learning techniques such as Conditional Random Fields (CRF) \cite{lafferty01}. However, for that, we need a large amount of training data for which a quadruplet $(x_{f},f,e,q)$ is available. In this work, we will use a corpus extended from \cite{besacier14} which  contains  6.7k utterances. We will investigate if this amount of data is enough to evaluate and test a joint model $p_{SLT}(q/x_{f},f,e)$.

As it is much easier to obtain data containing either the triplet $(x_{f},f,q)$ (automatically transcribed speech with manual references and  quality labels infered from word error rate estimation) or the triplet $(f,e,q)$ (automatically translated text with manual post-editions and  quality labels infered using tools such as TERpA \cite{snover08}) we can also recast the WCE problem with the following equation:

\begin{equation}
\label{eq:wce_asr_and_mt}
\hat{q}=\underset{q}{\operatorname{argmax} } \{p_{ASR}(q/x_{f},f)^\alpha*p_{MT}(q/e,f)^{1-\alpha}\}
\end{equation}

where $\alpha$ is a weight giving more or less importance to $WCE_{ASR}$ (quality assesment on transcription) compared to $WCE_{MT}$ (quality assesment on translation). It is important to note that $p_{ASR}(q/x_{f},f)$ corresponds to the quality estimation of the words in the target language based on features calculated on the source language (ASR). For that, what we do is projecting source quality scores to the target using word-alignment information between $e$ and $f$ sequences. This alternative approach (\textit{equation 6}) will be also evaluated in this work.

In both approaches -- \textit{joint} ($p_{SLT}(q/x_{f},f,e)$) and \textit{combined} ($p_{ASR}(q/x_{f},f)$ + $p_{MT}(q/e,f)$) -- some features need to be extracted from  ASR and MT modules. They are more precisely detailed in next subsections.

\subsection{WCE features for speech transcription (ASR)}
\label{sec:WCEasr}

In this work, we extract several types of features, which come from the ASR graph, from language model scores  and from a morphosyntactic analysis. These features are listed below (more details can be found in \cite{besacier14}):

\begin{itemize}
\item  Acoustic features: word duration (\textbf{F-dur}).
\item Graph features (extracted from the ASR word confusion networks): number of alternative (\textbf{F-alt}) paths between two nodes; word posterior probability (\textbf{F-post}). 
\item  Linguistic features (based on probabilities by the language model): word itself (\textbf{F-word}), 3-gram probability (\textbf{F-3g}), log probability (\textbf{F-log}), back-off level of the word (\textbf{F-back}), as proposed in \cite{JulienFayolle2010},
\item  Lexical Features: Part-Of-Speech (POS) of the word (\textbf{F-POS}),
\item  Context Features: Part-Of-Speech tags in the neighborhood of a given word (\textbf{F-context}).
\end{itemize}

For each word in the ASR hypothesis, we estimate the 9 features (F-Word; F-3g; F-back; F-log; F-alt; F-post; F-dur; F-POS; F-context) previously described.

In a preliminary experiment, we will evaluate these features for quality assessment in ASR only ($WCE_{ASR}$ task). Two different classifiers will be used: a variant of boosting classification algorithm called \textit{bonzaiboost} \cite{AntoineLaurent2014} (implementing the boosting algorithm \textit{Adaboost.MH} over deeper trees) and the Conditional Random Fields \cite{lafferty01}.




\subsection{WCE features for machine translation (MT)}
\label{sec:WCEmt}

A number of knowledge sources are employed for extracting features, in a total of 24 major feature types, see Table \ref{tab:all_features}. 

\begin{table*}[t]
\small
\centering
\caption{List of MT features extracted.} 
 \begin{tabular}{lll} 
\hline\noalign{\smallskip}
 1 Proper Name			& 10  Stop Word	& 19 WPP max 	\\
 2 Unknown Stem		& 11  Word context Alignments	& 20 Nodes 		\\
 3 Num. of Word Occ. 		& 12  POS context Alignments	& 21 Constituent Label  \\
 4 Num. of Stem Occ. & 13  Stem context Alignments & 22  Distance To Root   \\
 5 Polysemy Count -- Target		& 14 Longest Target $N$-gram Length	& 23  Numeric 			  \\
 6 Backoff Behaviour -- Target 		& 15 Longest Source $N$-gram Length	& 24 Punctuation  \\
 7 Alignment Features			& 16 WPP Exact 				&  \\
 8 Occur in Google Translate				& 17 WPP Any 				&   \\
 9  Occur in Bing Translator			& 18 WPP min  			&  \\


 \noalign{\smallskip}\hline
 \end{tabular}
\label{tab:all_features}
\end{table*}



It is important to note that we extract features regarding 
\textit{tokens} in the machine translation (MT) hypothesis sentence. In other words, 
one feature is extracted for each token in the MT output. 
So, in the Table \ref{tab:all_features}, \textit{target} refers to the feature coming from the MT 
hypothesis and
\textit{source} refers to a feature extracted from the source word aligned to 
the considered target word. More details on some of these features are given in 
the next subsections.

\subsubsection{Internal Features}
These features are given by the Machine Translation system, which outputs 
additional data like $N$-best list.


\textbf{Word Posterior Probability} (WPP) and \textbf{Nodes} features are extracted from a 
confusion network, which comes from the output of the machine translation 
$N$-best list. 
\textbf{WPP Exact} is the WPP value for each  word concerned at the exact same 
position in the graph. \textbf{WPP Any} extracts the same 
information at any position in the graph. \textbf{WPP Min} gives the smallest 
WPP value concerned by the transition and \textbf{WPP Max} its maximum. 

%
%
%


\subsubsection{External Features}
Below is the list of the external features used: 
\begin{itemize}
\item \textbf{Proper Name}: indicates if a word is a proper name (same binary features are extracted to know if a token is \textbf{Numerical}, \textbf{Punctuation} or \textbf{Stop Word}).

\item \textbf{Unknown Stem}: informs whether the stem of the considered word 
is known or not.

\item \textbf{Number of Word/Stem Occurrences}: counts the occurrences of a 
word/stem in the sentence.

\item \textbf{Alignment context features}: these features (\#11-13 in Table 
\ref{tab:all_features}) are based on collocations and proposed by \cite{bach11}.
Collocations could be an indicator 
for judging if a target word is generated 
by a particular source word. We also apply the reverse, the collocations 
regarding the source side (\#7 in Table 
\ref{tab:all_features} - simply called \textbf{Alignment Features}):

\renewcommand{\labelitemii}{$\bullet$}
\begin{itemize}
\item \emph{Source alignment context features}: the combinations of the target 
word, the source word (with which it is aligned), and one source word before 
and one source word after (left and right contexts, respectively).
\item \emph{Target alignment context features}: the combinations of the source 
word, the target word (with which it is aligned), and one target word before 
and one target word after.
\end{itemize}



 
 \item \textbf{Longest Target (or Source) $N$-gram Length}: 
 we seek to get the length ($n+1$) of the longest left sequence ($w_{i-n}$) concerned by the current 
 word ($w_i$) and known by the language model (LM) concerned (source and target sides).
 For example, 
 if the longest left sequence $w_{i-2},w_{i-1},w_i$ 
 appears in the target LM,
 the longest target n-gram value for $w_i$ will be 3. This value 
 ranges from 0 to the max order of the LM concerned. We also extract a redundant feature called \textbf{Backoff Behavior Target}.

%

\item The target word's constituent label (\textbf{Constituent Label}) and its depth 
in the constituent tree (\textbf{Distance to Root}) are extracted using 
a syntactic parser.  



\item \textbf{Target Polysemy Count}: we extract the polysemy count, 
which is the number of meanings of a word in a given language.


\item \textbf{Occurences in Google Translate} and \textbf{Occurences in Bing 
Translator}: 
in the translation hypothesis, we (optionally) test the presence of the target word in 
on-line translations given respectively by \textit{Google Translate} and 
\textit{Bing Translator}\footnote{Using this kind of feature is controversial, however we observed that such features are available in general use case scenarios, so we decided to include them in our experiments. Contrastive results without these 2 features will be also given later on.}.

\end{itemize}

A very similar feature set was used for a simple $WCE_{MT}$ task (English - Spanish MT, WMT 2013, 2014 quality estimation shared task) and obtained  very good performances \cite{luong13}. 
This preliminary experience in participating to the WCE shared task
in 2013 and 2014 lead us to the following observation: while feature processing 
is very important to achieve good performance, it requires to call a set of 
 heterogeneous NLP tools (for lexical, syntactic, semantic analyses). Thus, 
we recently proposed to unify the feature processing, together with the call of machine 
learning algorithms, in order to facilitate the design of confidence estimation 
systems. The open-source toolkit proposed (written in \textit{Python} and made 
available on \textit{github}\footnote{\url{http://github.com/besacier/WCE-LIG}}) integrates some standard as well as in-house 
features that have proven useful for WCE (based on our experience in WMT 2013 
and 2014). 

In this paper, we will use only Conditional Random Fields \cite{lafferty01} (CRFs) as our machine learning method, with WAPITI toolkit \cite{lavergne10}, to train our WCE estimator based on MT features.

%
%

%

\section{Experimental setup}
\label{sec:db}

\subsection{Dataset}
\label{subsec:dataset}

\subsubsection{Starting point: an existing MT Post-edition corpus}
\label{subsubsec:starting_point}

For a French-English translation task, we used our SMT system to obtain the translation hypothesis for 10,881 source sentences taken from news corpora of the WMT (Workshop on Machine Translation) evaluation campaign (from 2006 to 2010).  Post-editions were obtained from non professional translators using a crowdsourcing platform. More details on the baseline SMT system used can be found in \cite{potet10} and more details on the post-edited corpus can be found in \cite{potet12}.
It is worth mentionning, however, that a sub-set (311 sentences) of these collected post-editions was assessed by a professional translator and 87.1\% of post-editions were judged to improve the hypothesis  

Then, the word label setting for WCE was done using TERp-A toolkit \cite{snover08}. Table \ref{tab:Terpa} illustrates the labels generated by TERp-A for one hypothesis and post-edition pair. Each word or phrase in the hypothesis is aligned to a word or phrase in the post-edition with different types of edit: “I” (insertions), “S” (substitutions), “T” (stem matches), “Y” (synonym matches), and “P” (phrasal substitutions). The lack of a symbol indicates an exact match and will be replaced by “E” thereafter.  We do not consider the words marked with “D” (deletions) since they appear only in the reference. However, later on, we will have to train binary classifiers ($good$/$bad$) so we re-categorize the obtained 6-label set into binary set: The E, T and Y belong to the $good$ (G), whereas the S, P and I belong to the $bad$ (B) category.

\begin{table}
\caption{Example of training label obtained using TERp-A.}
\label{tab:Terpa}       
\begin{tabular}{lllllll}
\hline\noalign{\smallskip}
\bf Reference & The & consequence & of & the & fundamentalist\\
\noalign{\smallskip}\hline\noalign{\smallskip}
&&S&&&S\\
\noalign{\smallskip}\hline
\bf Hyp After Shift & The & result & of & the & hard-line\\
\noalign{\smallskip}\hline
\noalign{\smallskip}\hline
\bf Reference & movement &&also& has & its importance &.\\
\noalign{\smallskip}\hline\noalign{\smallskip}
&Y&I && D&P&\\
\noalign{\smallskip}\hline
\bf Hyp After Shift & trend & is&also& &important &.\\
\noalign{\smallskip}\hline
\end{tabular}
\end{table}

\subsubsection{Extending the corpus with speech recordings and transcripts}
\label{subsubsec:augmenting_corpus}

The \textit{dev} set and \textit{tst} set of this corpus were recorded by french native speakers.
Each sentence was uttered by 3 speakers, leading to 2643 and 4050 speech recordings for \textit{dev} set and \textit{tst} set, respectively. 
For each speech utterance, a 
quintuplet containing: ASR output ($f_{hyp}$), verbatim transcript ($f_{ref}$), 
English text translation output ($e_{hyp_{mt}}$), speech translation output 
($e_{hyp_{slt}}$) and post-edition of translation ($e_{ref}$), was made 
available. 
This corpus is available on a \textit{github} 
repository\footnote{https://github.com/besacier/WCE-SLT-LIG/}. More details are given in  table \ref{tab:corpus_speech_recordings}. The total length of the \textit{dev} and \textit{tst} speech corpus obtained are 16h52, since some utterances were pretty long. 


\begin{table}[!ht]
\caption{Details on our \textit{dev} and \textit{test} corpora for SLT.}
\label{tab:corpus_speech_recordings}
\begin{tabular}{lllll}
\hline\noalign{\smallskip}
\textbf{Corpus } & \textbf{\#sentences} & \textbf{\#speech recordings} & \textbf{\#speakers}             & \textbf{Duration} \\
\noalign{\smallskip}\hline\noalign{\smallskip}
\textit{dev} & 881         & 2643                & 15 (9 women + 6 men)   & 5h51     \\
\textit{tst} & 1350        & 4050                & 27 (11 women + 16 men) & 11h01    \\
\noalign{\smallskip}\hline
\end{tabular}
\end{table}

\subsection{ASR Systems}
\label{subsec:asr_system}

To obtain the speech transcripts ($f_{hyp}$), we built a French ASR system based on KALDI toolkit \cite{Povey_ASRU2011}. 
Acoustic models are trained using several corpora (ESTER, REPERE, ETAPE and BREF120) representing more than 600 hours of french transcribed speech. 

The baseline GMM system is based on mel-frequency cepstral coefficient (MFCC) acoustic features (13 coefficients expanded with delta and double delta features and energy : 40 features) with various feature transformations including linear discriminant analysis (LDA), maximum likelihood linear transformation (MLLT), and feature space maximum likelihood linear regression (fMLLR) with speaker adaptive training (SAT). The GMM acoustic model makes initial phoneme alignments of the training data set for the following DNN acoustic model training. 

The speech transcription process is carried out in two passes: an automatic transcript is generated with a GMM-HMM model of 43182 states and 250000 Gaussians. Then word graphs outputs obtained during the first pass are used to compute a fMLLR-SAT transform on each speaker. The second pass is performed using DNN acoustic model trained on acoustic features normalized with the fMLLR matrix.

CD-DNN-HMM acoustic models are trained (43 182 context-dependent states) using GMM-HMM topology. 

We propose to use two 3-gram language models trained on French ESTER corpus \cite{Galliano06corpusdescription} as well as on French Gigaword (vocabulary size are respectively 62k and 95k). The ASR systems LM weight parameters are tuned through WER on the \textit{dev} corpus. Details on these two language models can be found in table 4.

In our experiments we propose two ASR systems based on the previously described language models. The first system ($ASR1$) uses the small language model allowing a fast ASR system (about 2x Real Time), while in the second system lattices are rescored with a big language model (about 10x Real Time) during a third pass. 

\begin{table}
\caption{Details on language models (LM) used in our two ASR systems.}
\label{tab:LM}
\begin{tabular}{llcl}
\hline
\textbf{LM }   & \textbf{1-gram} & \textbf{2-grams} & \textbf{3-grams} \\
\hline
small ($ASR1$) & 62K     & 1M  & 59M  \\
\hline
big ($ASR2$)  & 95K     & 49M & 301M \\
\hline
\end{tabular}
\end{table}

\textit{Table \ref{tab:asr_performances}} presents the performances obtained by two above ASR systems.

\begin{table}[!ht]
\caption{ASR performance (WER) on our \textit{dev} and \textit{test} set for the two different ASR systems}
\label{tab:asr_performances}       
\begin{tabular}{lll}
\hline\noalign{\smallskip}
\bf Task &  \textbf{\textit{dev} set} & \textbf{\textit{tst} set}\\
\noalign{\smallskip}\hline\noalign{\smallskip}
\textit{ASR1} & 21.86\% & 17.37\% \\
\textit{ASR2} & 16.90\% & 12.50\% \\
\noalign{\smallskip}\hline
\end{tabular}
\end{table}

These WER may appear as rather high according to the task (transcribing read news). 
A deeper analysis shows that these news contain  a lot of 
foreign named entities, especially in our \textit{dev} set. This part of the 
data is extracted from French medias 
dealing with european economy in EU. This could also explain why the 
scores are significantly different between \textit{dev} and \textit{test} sets.
In addition, automatic post-processing is applied to ASR output in order to 
match requirements of standard input for machine translation.

\subsection{SMT System}
\label{subsec:smt_system}
We used \textit{moses} phrase-based translation toolkit \cite{koehn07} to translate French ASR into English ($e_{hyp}$).
This medium-size system was trained using a subset of data provided for IWSLT 2012 evaluation  \cite{iwslt2012_campaign}: Europarl, Ted and News-Commentary corpora.
The total amount is about 60M words. 
We used an adapted target language model trained on specific data (News Crawled 
corpora) similar to our evaluation corpus (see \cite{potet10}). This standard SMT system will be used in all experiments reported in this paper.

\subsection{Obtaining quality assessment labels for SLT}
\label{subsec:obtaining_label}


After building an ASR system, we  have a new element of our desired quintuplet: the ASR output $f_{hyp}$. It is the noisy version of our already available verbatim transcripts called $f_{ref}$. 
This ASR output ($f_{hyp}$) is then translated by the exact same SMT system \cite{potet10} already mentionned in subsection \ref{subsec:smt_system}.
This new output translation is called $e_{hyp_{slt}}$ and it is a degraded version of $e_{hyp_{mt}}$ (translation of $f_{ref}$). 

At this point, a strong assumption we made has to be revealed: we re-used the post-editions obtained from the text translation task (called $e_{ref}$), to infer the quality (G, B) labels of our speech translation output $e_{hyp_{slt}}$. The word label setting for WCE is also done using TERp-A toolkit \cite{snover08} between $e_{hyp_{slt}}$ and $e_{ref}$. This assumption, and the fact that initial MT post-edition can be also used to infer labels of a SLT task, is reasonnable regarding results (later presented in table \ref{tab:mt-slt-res-dev-set} and table \ref{tab:mt-slt-res-tst-set}) where it is shown that there is not a huge difference between the MT and SLT performance (evaluated with BLEU).

The remark above is important and this is what makes the value of this corpus. For instance, other corpora such as the TED corpus compiled by LIUM\footnote{http://www-lium.univ-lemans.fr/fr/content/corpus-ted-lium} contain also a quintuplet with ASR output, verbatim transcript, MT output, SLT output  and target translation. But there are 2 main differences: first, the target translation is a manual translation of the prior subtitles so this is not a post-edition of an automatic translation (and we have no guarantee that the $good$/$bad$ labels extracted from this will be reliable for WCE training and testing); secondly, in our corpus, each sentence is uttered by 3 different speakers which introduces speaker variability in the database and allows us to deal with different ASR outputs for a single source sentence.

\subsection{Final corpus statistics}
The final corpus obtained is summarized in table \ref{tab:summary-db}, where we also clarify how the WCE labels were obtained. For the test set, we now have all the data needed to evaluate WCE for 3 tasks: 
\begin{itemize}
 \item \textbf{ASR}: extract $good$/$bad$ labels by calculating WER between $f_{hyp}$ and $f_{ref}$,
 \item \textbf{MT}: extract $good$/$bad$ labels by calculating TERp-A between $e_{hyp_{mt}}$ and $e_{ref}$,
 \item \textbf{SLT}: extract $good$/$bad$ labels by calculating TERp-A between $e_{hyp_{slt}}$ and $e_{ref}$.
 \end{itemize}


\begin{table}[!ht]
\caption{Overview of our post-edition corpus for SLT.}
\label{tab:summary-db}       
\begin{tabular}{llll}
\hline\noalign{\smallskip}
\bf Data & \bf \# \textit{dev} utt & \bf \# \textit{test} utt & \bf method to obtain WCE labels\\
\noalign{\smallskip}\hline\noalign{\smallskip}
$f_{ref}$ & 881 & 1350 & \\
$f_{hyp}$ & 881*3 & 1350*3 & wer($f_{hyp}$, $f_{ref}$)\\
\hline
$e_{hyp_{mt}}$ & 881 & 1350 & terpa($e_{hyp_{mt}}$, $e_{ref}$) \\
$e_{hyp_{slt}}$ & 881*3 & 1350*3 & terpa($e_{hyp_{slt}}$, $e_{ref}$) \\
$e_{ref}$ & 881 & 1350 &  \\
\noalign{\smallskip}\hline
\end{tabular}
\end{table}

Table \ref{tab:ex} gives an example of the quintuplet available in our corpus. One transcript ($f_{hyp1}$) has 1 error while the other one ($f_{hyp2}$) has 4. This leads to respectively 2 B labels ($e_{hyp_{slt1}}$) and 4 B labels ($e_{hyp_{slt2}}$) in the speech translation output, while $e_{hyp_{mt}}$ has only one B label.

\begin{table}[h]
\caption{Example of quintuplet with associated labels.}
\begin{tabular}{lllll}
\noalign{\smallskip}\hline
$f_{ref}$ & quand & notre  &  cerveau  & chauffe \\
\noalign{\smallskip}\hline
$f_{hyp1}$ & \textit{comme} & notre  &  cerveau  & chauffe \\
labels ASR & B & G & G & G\\
$f_{hyp2}$ & \textit{qu'} & \textit{entre}  &  \textit{serbes}  & \textit{au} chauffe \\
labels ASR & B & B & B & B   G\\
\noalign{\smallskip}\hline
$e_{hyp_{mt}}$ & when & our  &  brains  & \textit{chauffe} \\
labels MT & G & G & G & B\\
\noalign{\smallskip}\hline
$e_{hyp_{slt1}}$ & \textit{as} & our  &  brains  & \textit{chauffe} \\
labels SLT & B & G & G & B\\
$e_{hyp_{slt2}}$ & \textit{between} &  \textit{serbs}   & \textit{in}  & \textit{chauffe} \\
labels SLT & B & B & B & B\\
\noalign{\smallskip}\hline
$e_{ref}$ & when & our  &  brain  & heats up \\
\noalign{\smallskip}\hline
\end{tabular}
\label{tab:ex}
\end{table}

Table \ref{tab:mt-slt-res-dev-set} and table \ref{tab:mt-slt-res-tst-set} summarize baseline ASR, MT and SLT performances obtained on our corpora, as well as the distribution of good (G) and bad (B) labels inferred for both tasks. Logically, the percentage of (B) labels increases from MT to SLT task in the same conditions.


\begin{table}[!ht]
\caption{MT and SLT performances on our \textit{dev} set.}
\begin{tabular}{lllll}
\noalign{\smallskip}\hline
\bf Task & \bf ASR (WER) & \bf MT (BLEU) & \bf \% G (good) & \bf \% B (bad)\\
\noalign{\smallskip}\hline
MT & 0\% & 49.13\%  &  76.93\%  & 23.07\% \\
\noalign{\smallskip}\hline
SLT (ASR1) & 21.86\% & 26.73\% & 62.03\%  & 37.97\% \\
\noalign{\smallskip}\hline
SLT (ASR2) & 16.90\% & 28.89\% & 63.87\%  & 36.13\% \\
\noalign{\smallskip}\hline
\end{tabular}
\label{tab:mt-slt-res-dev-set}
\end{table}

\begin{table}[h]
\caption{MT and SLT performances on our \textit{tst} set.}
\begin{tabular}{lllll}
\noalign{\smallskip}\hline
\bf Task & \bf ASR (WER) & \bf MT (BLEU) & \bf \% G (good) & \bf \% B (bad)\\
\noalign{\smallskip}\hline
MT & 0\% & 57.87\%  &  81.58\%  & 18.42\% \\
\noalign{\smallskip}\hline
SLT ($ASR1$) & 17.37\% & 30.89\% & 61.12\%  & 38.88\% \\
\noalign{\smallskip}\hline
SLT ($ASR2$) & 12.50\% & 33.14\% & 62.77\%  & 37.23\% \\
\noalign{\smallskip}\hline
\end{tabular}
\label{tab:mt-slt-res-tst-set}
\end{table}

\section{Experiments on WCE for SLT}
\label{sec:WCEjoint}

\subsection{SLT quality assessment using only MT or ASR features}

We first report in Table \ref{tab:wce_performance_diff_feat_for_tst_set} the baseline WCE results obtained using MT or ASR features separately. 
In short, we evaluate the performance of 4 WCE systems for different tasks:
\begin{itemize}
\item The first and second systems (WCE for ASR / ASR feat.) use ASR features described in section ~\ref{sec:WCEasr}  with two different classifiers (CRF or Boosting).
\item The third system (WCE for SLT / MT feat.) uses only MT features described in section ~\ref{sec:WCEmt} with CRF classifier.
\item The fourth system (WCE for SLT / ASR feat.) uses only ASR features described in section ~\ref{sec:WCEasr}  with CRF classifier  (so this is predicting SLT output confidence using only ASR confidence features!). Word alignment information between $f_{hyp}$ and  $e_{hyp}$ is used to project the WCE scores coming from ASR, to the SLT output,
\end{itemize}

In all experiments reported in this paper, we evaluate the performance of our classifiers by using the average between the F-measure for $good$ labels and the F-measure for $bad$ labels that are calculated by the common evaluation metrics: Precision, Recall and F-measure for $good$/$bad$ labels. Since two ASR systems are available, \textit{F-mes1} is obtained for SLT based on $ASR1$ whereas \textit{F-mes2} is obtained for SLT based on $ASR2$.  For the results of Table \ref{tab:wce_performance_diff_feat_for_tst_set}, the classifier is evaluated on the \textit{tst} part of our corpus and trained on the \textit{dev} part. 



\begin{table}[!ht]
\caption{WCE performance with different feature sets for \textit{tst} set  (training is made on \textit{dev} set) - *for MT feat, removing \textit{OccurInGoogleTranslate} and  \textit{OccurInBingTranslate } features lead to 59.40\% and 58.11\% for  \textit{F-mes1} and \textit{F-mes2} respectively}
\label{tab:wce_performance_diff_feat_for_tst_set}       
\begin{tabular}{llllll}
\hline\noalign{\smallskip}
\bf task   & \bf WCE for ASR  & \bf WCE for ASR  & \bf WCE  for SLT  & \bf WCE for SLT\\
feat. type &  ASR feat.       &  ASR feat.       & MT feat.          & ASR feat.       \\
           &  $p(q/x_{f},f)$  &  $p(q/x_{f},f)$   & $p(q/f,e)$       & $p(q/x_{f},f)$  \\
           & (CRFs)           & (Boosting)    &                  & projected to $e$\\
\noalign{\smallskip}\hline\noalign{\smallskip}
\textit{F-mes1} & 68.71\% & 64.27\%  & 60.55\%* & 49.67\% \\
\textit{F-mes2} & 59.83\% & 62.61\%  & 59.83\%* & 44.56\% \\
\noalign{\smallskip}\hline
\end{tabular}
\end{table}

Concerning WCE for ASR, we observe that Fmeasure decreases when ASR WER is lower (\textit{F-mes2}$<$\textit{F-mes1} while $WER_{ASR2}<WER_{ASR1}$). 
So quality assessment in ASR seems to become harder as the ASR system improves. This could be due to the fact that the ASR1 errors recovered by bigger LM in ASR2 system were easier to detect. Anyway, this conclusion should be considered with caution since both results (\textit{F-mes1} and \textit{F-mes2}) are not directly comparable because they are evaluated on different references (proportion of  $good$/$bad$ labels differ as ASR system differ). The effect of the classifier (CRF or Boosting) is not conclusive since CRF is better for \textit{F-mes1} and worse for \textit{F-mes2}. Anyway, we decide to use CRF for all our future  experiments since this is the classifier integrated in WCE-LIG \cite{servan-toolkit-2015} toolkit. 

Concerning WCE for SLT, we observe that Fmeasure is better using MT features rather than ASR features (quality assessment for SLT more dependent of MT features than ASR features). Again, Fmeasure decreases when ASR WER is lower (\textit{F-mes2}$<$\textit{F-mes1} while $WER_{ASR2}<WER_{ASR1}$).  For MT features, removing \textit{OccurInGoogleTranslate} and  \textit{OccurInBingTranslate } features lead to 59.40\% and 58.11\% for  \textit{F-mes1} and \textit{F-mes2} respectively.

In the next subsection, we try to see if the use of both MT and ASR features improves quality assessment for SLT.

\subsection{SLT quality assessment using both MT and ASR features}

We now report in Table \ref{tab:wce_performance_joint_feat_for_tst_set} WCE for SLT results obtained using both MT and ASR features. More precisely we evaluate two different approaches (\textit{combination} and \textit{joint}):
\begin{itemize}
\item The first system (WCE for SLT / MT+ASR feat.) combines the output of two separate classifiers based on ASR and MT features. In this approach, ASR-based confidence score of the source is projected to the target SLT output and combined with the MT-based confidence score as shown in \textit{equation 6} 
(we did not tune the $\alpha$ coefficient and set it \textit{a priori} to 0.5).

\item The second system (joint feat.) trains a single WCE system for SLT (evaluating $p(q/x_{f},f,e)$ as in \textit{equation 5} using joint ASR features and MT features. All ASR features are projected to the target words using automatic word alignments. However, a problem occur when a target word does not have any source word aligned to it. In this case, we decide to duplicate the ASR features of its previous target word. Another problem occur when a target word is aligned to more than one source word. In that case, there are several strategies to infer the 9 ASR features: average or max over numerical values, selection or concatenation over symbolic values (for F-word and F-POS), etc. 
Three different variants of these strategies (shown in Table \ref{tab:example_asr_features_joint_feature_vector}) are evaluated here.

\end{itemize}


\begin{table}[!ht]
\caption{Different strategies to project ASR features to a target word when it is aligned to more than one source word.           
*It should be noted that \textbf{F-context} features are the combinations of the source word (\textbf{F-word}) and one POS of source word (\textbf{F-POS}) before and one POS of source word (\textbf{F-POS}) after. 
}
\label{tab:example_asr_features_joint_feature_vector}
\begin{tabular}{llll}
\hline\noalign{\smallskip}
\textbf{\textit{ASR} Feat} & \textbf{Joint 1}  & \textbf{Joint 2} & \textbf{Joint 3}          \\
\noalign{\smallskip}\hline\noalign{\smallskip}
F-post        & avg(F-post1, F-post2) & avg(F-post1, F-post2) & avg(F-post1, F-post2) \\
F-log         & avg(F-log1, F-log2)   & avg(F-log1, F-log2)   & avg(F-log1, F-log2)   \\
F-back        & avg(F-back1, F-back2) & avg(F-back1, F-back2) & avg(F-back1, F-back2) \\
F-dur         & max(F-dur1, F-dur2)   & max(F-dur1, F-dur2)   & max(F-dur1, F-dur2)   \\
F-3g          & max(F-3g1, F-3g2)   & max(F-3g1, F-3g2)   & max(F-3g1, F-3g2)   \\
F-alt         & max(F-alt1, F-alt2)   & max(F-alt1, F-alt2)   & max(F-alt1, F-alt2)   \\
F-word        & F-word1               & F-word2               & F-word1\_F-word2      \\
F-POS         & F-POS1                & F-POS2                & F-POS1\_F-POS2        \\
F-context  & F-context*          & F-context*          & F-context* \\
\noalign{\smallskip}\hline
\end{tabular}
\end{table}



 

\begin{table}[!ht]
\caption{WCE performance with combination (MT+ASR) or joint (MT,ASR) feature sets for \textit{tst} set (training is made on \textit{dev} set) - * For \textit{Joint 1} feat, removing \textit{OccurInGoogleTranslate} and  \textit{OccurInBingTranslate } features lead to 59.14\% and 57.75\% for  \textit{F-mes1} and \textit{F-mes2} respectively.
}
\label{tab:wce_performance_joint_feat_for_tst_set}       
\begin{tabular}{lllll}
\hline\noalign{\smallskip}
\bf task   & \bf WCE for SLT& \bf WCE for SLT& \bf WCE for SLT& \bf WCE for SLT\\
feat. type & MT+ASR feat.   &Joint feat. 1   &Joint feat. 2   &Joint feat. 3\\
                &     $p_{ASR}(q/x_{f},f)^\alpha$     &  $p(q/x_{f},f,e)$  &  $p(q/x_{f},f,e)$   & $p(q/x_{f},f,e)$       \\
                &     $*p_{MT}(q/e,f)^{1-\alpha}$     &    &    &       \\

\noalign{\smallskip}\hline\noalign{\smallskip}
\textit{F-mes1} & 52.99\%&\textbf{60.29\%*}&60.17\%&60.23\%\\
\textit{F-mes2} & 48.46\%&\textbf{59.23\%*}&59.20\%&58.99\%\\
\noalign{\smallskip}\hline
\end{tabular}
\end{table}

The results of Table \ref{tab:wce_performance_joint_feat_for_tst_set} show that joint ASR and MT features do not improve WCE performance: \textit{F-mes1} and \textit{F-mes2} are slightly worse than  those of table 9 (WCE for SLT / MT features only). We also observe that simple combination (MT+ASR) degrades the WCE performance. This latter observation may be due to different behaviors of $WCE_{MT}$ and $WCE_{ASR}$ classifiers which makes the weighted combination ineffective. Moreover, the disappointing performance of our joint classifier may be due to an insufficient training set (only 2683 utterances in \textit{dev}!). Finally, removing \textit{OccurInGoogleTranslate} and  \textit{OccurInBingTranslate } features for \textit{Joint} lowered \textit{F-mes} between 1\% and 1.5\%.

These observations lead us to investigate the behaviour of our WCE approaches for a large range of $good$/$bad$ decision threshold and with a new protocol where we reverse \textit{dev} and \textit{tst}. So, in the next experiments of this subsection, we will report WCE evaluation results obtained on \textit{dev} (2683 utt.) with classifiers trained on \textit{tst} (4050 utt.). Finally, the different strategies used to project ASR features when a target word is aligned to more than one source word do not lead to very different performance: we will use strategy \textit{joint 1} in the future.

\begin{figure*}
  \includegraphics[width=0.75\textwidth]{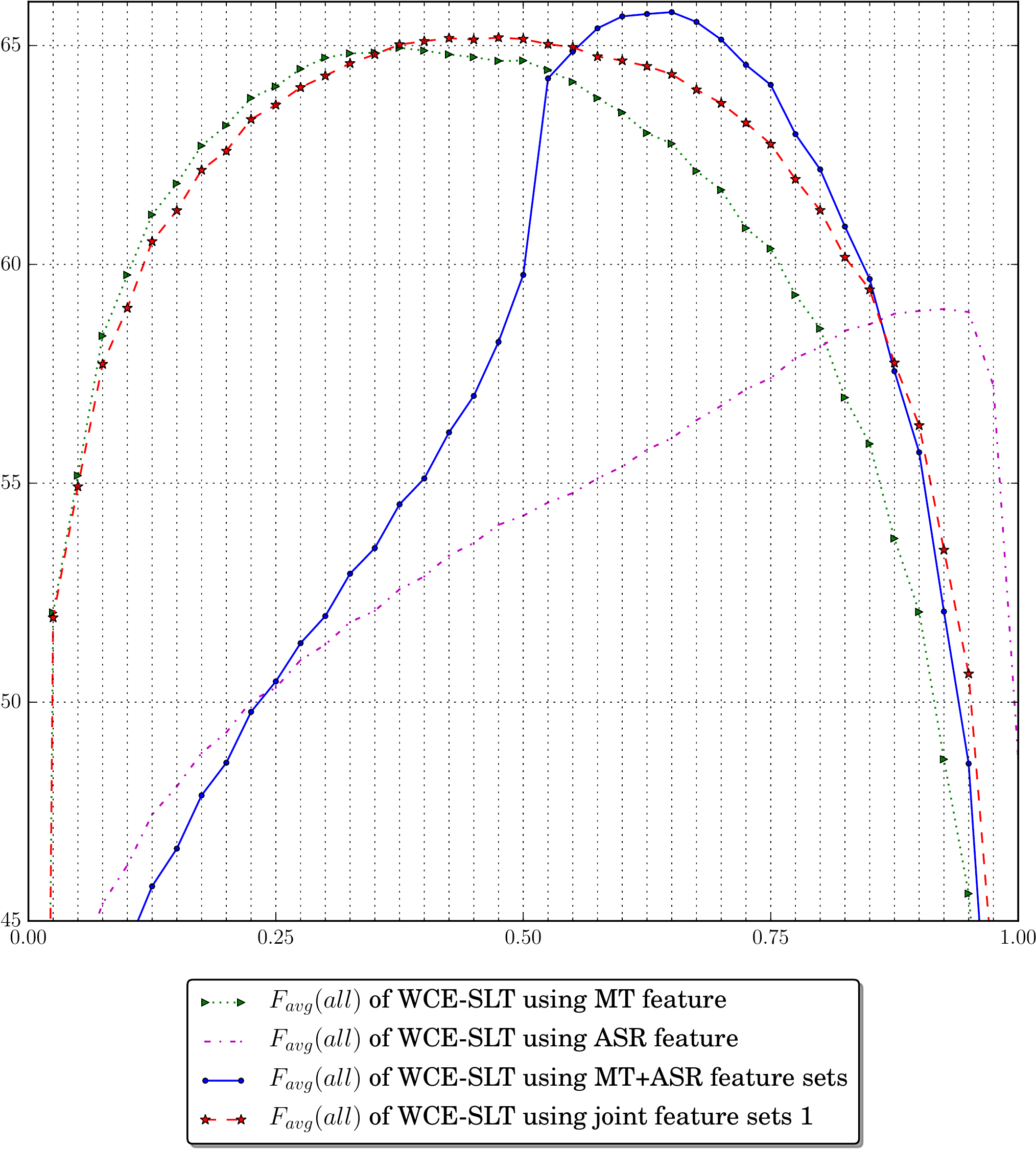}
\caption{Evolution of system performance (y-axis - \textit{F-mes1} - ASR1) for \textit{dev} corpus (2683 utt) along decision threshold variation (x-axis) - training is made on 4050 utt.}
\label{fig:WCE_SLT_for_MT_ASR_feat_AND_Joint_feat_for_TestDEV_corpus_all_curves1}       
\end{figure*}

\begin{figure*}
  \includegraphics[width=0.75\textwidth]{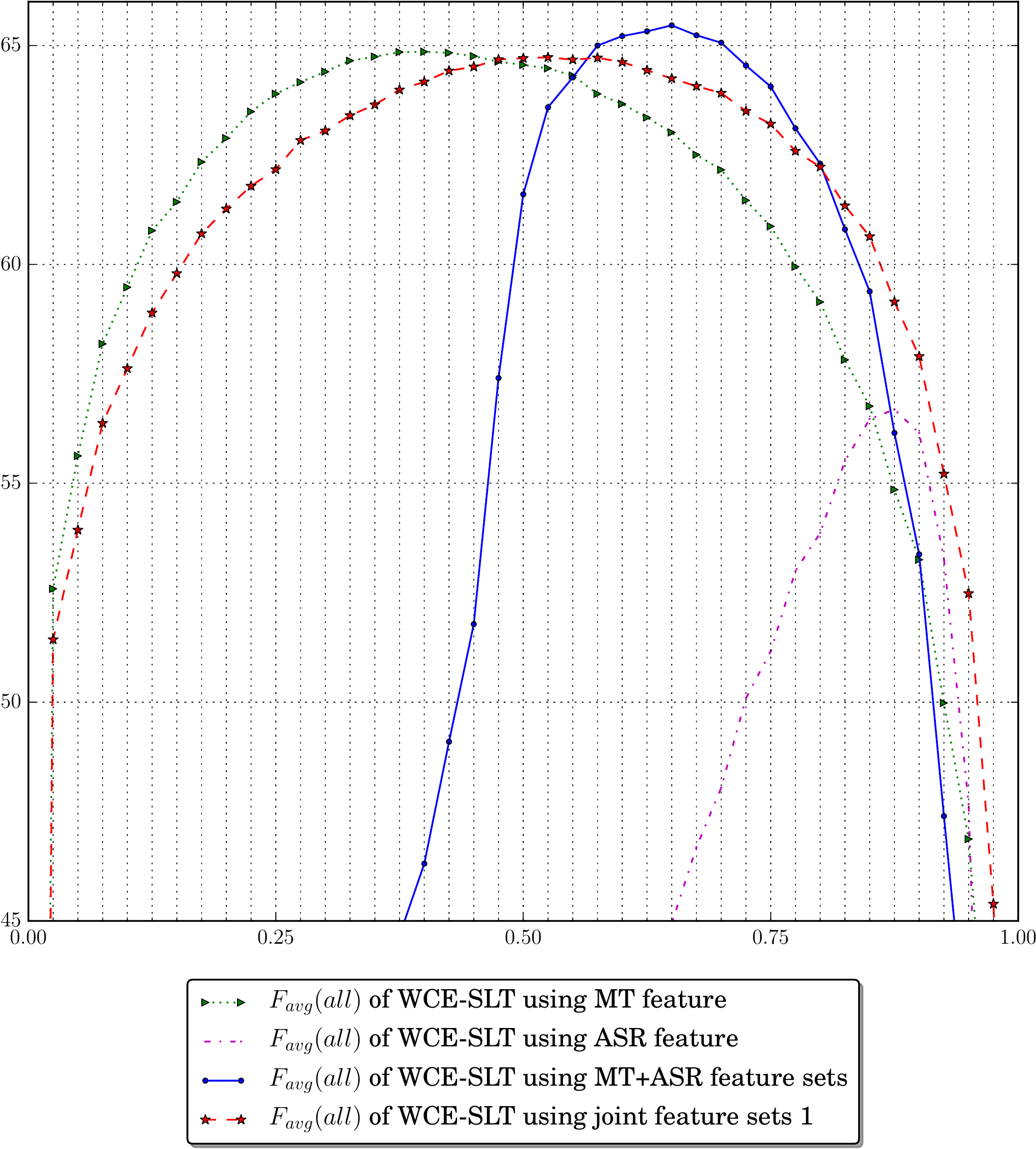}
\caption{Evolution of system performance (y-axis - \textit{F-mes2} - ASR2) for \textit{dev} corpus (2683 utt) along decision threshold variation (x-axis) - training is made on 4050 utt.}
\label{fig:WCE_SLT_for_MT_ASR_feat_AND_Joint_feat_for_TestDEV_corpus_all_curves2}       
\end{figure*}

While the previous tables provided WCE performance for a single point of interest ($good$/$bad$ decision threshold set to 0.5), the curves of figures \ref{fig:WCE_SLT_for_MT_ASR_feat_AND_Joint_feat_for_TestDEV_corpus_all_curves1} and \ref{fig:WCE_SLT_for_MT_ASR_feat_AND_Joint_feat_for_TestDEV_corpus_all_curves2} show the full picture of our WCE systems (for SLT) using speech transcriptions systems $ASR1$ and $ASR2$, respectively. We observe that the classifier based on ASR features has a very different behaviour than the classifier based on MT features which explains why their simple combination (MT+ASR) does not work very well for the default decision threshold (0.5). However, for threshold above 0.5, the use of both ASR and MT features is beneficial. This is interesting because higher thresholds improves the Fmeasure on $bad$ labels (so improves error detection). Both curves are similar whatever the ASR system used. These results suggest that with enough development data for appropriate threshold tuning (which we do not have for this very new task), the use of both ASR and MT features should improve error detection in speech translation (blue and red curves are above the green curve for higher decision threshold\footnote{Corresponding to optimization of the Fmeasure on $bad$ labels (errors)}). We also analyzed the $Fmeasure$ curves for $bad$ and $good$ labels separately\footnote{Not reported here due to space constraints.}: if we consider, for instance $ASR1$ system, for decision threshold equals to 0.75, the Fmeasure on $bad$ labels is equivalent (60\%) for 3 systems (\textit{Joint, MT+ASR} and \textit{MT}) while the Fmeasure on $good$ labels is 61\% when using \textit{MT} features only, 66\% when using \textit{Joint} features and 68\% when using \textit{MT+ASR} features. In other words, for a fixed performance on $bad$ labels, the Fmeasure on $good$ labels is improved using all information available (ASR and MT features). Finally, if we focus on \textit{Joint} versus \textit{MT+ASR}, we notice that the range of the threshold where performance are stable is larger for \textit{Joint} than for \textit{MT+ASR}.

\section{Feature Selection}
\label{subsec:featSel}

In this  section, we try to better understand the contribution of each (ASR or MT) feature by applying feature selection on our joint WCE classifier. In these experiments, we decide to keep  \textit{OccurInGoogleTranslate} and  \textit{OccurInBingTranslate} features. 

We choose the Sequential Backward Selection (SBS) algorithm which is a top-down algorithm starting  from a feature
set noted $Y_k$ (which denotes the set of all features) and sequentially removing the most irrelevant one ($x$) that maximizes the Mean F-Measure, $MF(Y_k-x)$. In our work, we examine until the set $Y_k$ contains only one remaining feature. 
Algorithm 1 
summarizes the whole process.

\begin{algorithm}
\label{algo:SBS}
\caption{Sequential Backward Selection (SBS) algorithm for feature selection. $Y_k$ denotes the set of all features and $x$ is the feature removed at each step of the algorithm
}
\begin{algorithmic}
\While {size of $Y_k$ $>$ $0$ }

\State $maxval = 0$ 
\For{$x$ $\in$ $Y_k$}

\If {$maxval$ $<$ $MF(Y_k-x)$}
    \State $maxval\gets MF(Y_k-x)$
    \State $worstfeat\gets x$
\EndIf

\EndFor

\State remove $worstfeat$ from $Y_k$

\EndWhile
\end{algorithmic}
\end{algorithm}

The results of the SBS algorithm can be found in table \ref{tab:rank_feature_selection_SBS_for_dev_corpus_tgt_slt1} which ranks all  joint features used in WCE for SLT  by order of importance after applying the algorithm on \textit{dev}. We can see that the SBS algorithm is not very stable and is clearly influenced by the ASR system ($ASR1$ or $ASR2$) considered in SLT. Anyway, if we focus on the features that are in the top-10 best in both cases, we find that the most relevant ones are:
\begin{itemize}
 \item \textit{Occur in Google Translate} and \textit{Occur in Bing Translate} (diagnostic from other MT systems),
 \item \textit{Longest Source N-gram Length, Target Backoff Behaviour} (source or target N-gram features)
 \item \textit{Stem Context Alignment} (source-target alignment feature)
 \end{itemize}
 
We also observe that the most relevant ASR features (in bold in table \ref{tab:rank_feature_selection_SBS_for_dev_corpus_tgt_slt1}) are \textit{F-3g, F-POS} and \textit{F-back} (lexical and linguistic features) whereas ASR acoustic and graph based features are among the worst (\textit{F-post, F-alt, F-dur}). So, in our experimental setting, it seems that MT features are more influent than ASR features. Another surprising result is the relatively low rank of word posterior probability (WPP) features whereas we were expecting to see them among the top features (as shown in \cite{luong15} where \textit{WPP Any} is among the best features for WCE in MT).

\begin{table}
\centering
\caption{Rank of each feature according to the Sequential Backward Selection algorithm - WCE for SLT task - Joint (ASR,MT) features used - Feature selection applied on \textit{dev} corpus for both \textit{ASR1} and  \textit{ASR2}  - ASR features are in bold.}
\label{tab:rank_feature_selection_SBS_for_dev_corpus_tgt_slt1}
\begin{tabular}{ccl|ccl}
\hline\noalign{\smallskip}
\textbf{Rank} & \textbf{Rank} & \textbf{Feature} & \textbf{Rank} & \textbf{Rank} & \textbf{Feature} \\
\textbf{\textit{ASR1}} & \textbf{\textit{ASR2}} &  &   \textbf{\textit{ASR1}} & \textbf{\textit{ASR2}} & \\
\noalign{\smallskip}\hline\noalign{\smallskip}
 1 & 4 & Occur in Google Translate & 18 & 14 & Numeric \\
 2 & 2 & Longest Source $N$-gram Length & 19 & 30 &  Proper Name \\ 
 3 & 5 & Target Backoff Behaviour & 20 & 20 & Unknown Stem \\ 
 4 & 22 & Constituent Label & 21 & 24 & Number of Word Occurrences \\ 
 5 & 1 & Occur in Bing Translate & 22 & 23 & \textbf{F-alt} \\ 
 6 & 11 & \textbf{F-3g} & 23 & 15 & Nodes \\ 
 7 & 16 & WPP Exact & 24 & 8 & \textbf{F-log} \\ 
 8 & 7 & Stem Context Alignment & 25 & 32 &  \textbf{F-context} \\ 
 9 & 17 & WPP Max & 26 & 19 &  Longest Target $N$-gram Length \\ 
10 & 12 & Number of Stem Occurrences & 27 & 6 &  WPP Any \\ 
11 & 21 & Polysemy Count - Target & 28 & 29 & POS Context Alignment \\ 
12 & 3 & \textbf{F-POS} & 29 & 10 &  \textbf{F-post} \\
13 & 18 & Stop Word & 30 &  28 & Word Context Alignment \\ 
14 & 25 & Distance to Root & 31 & 31 & \textbf{F-dur} \\
15 & 13 & \textbf{F-back} & 32 & 9 &  Alignment Features \\
16 & 26 & WPP Min & 33 & 33 & \textbf{F-word} \\
17 & 27 & Punctuation &  & &  \\
\noalign{\smallskip}\hline
\end{tabular}
\end{table}

%

\begin{figure*}
  \includegraphics[width=0.75\textwidth]{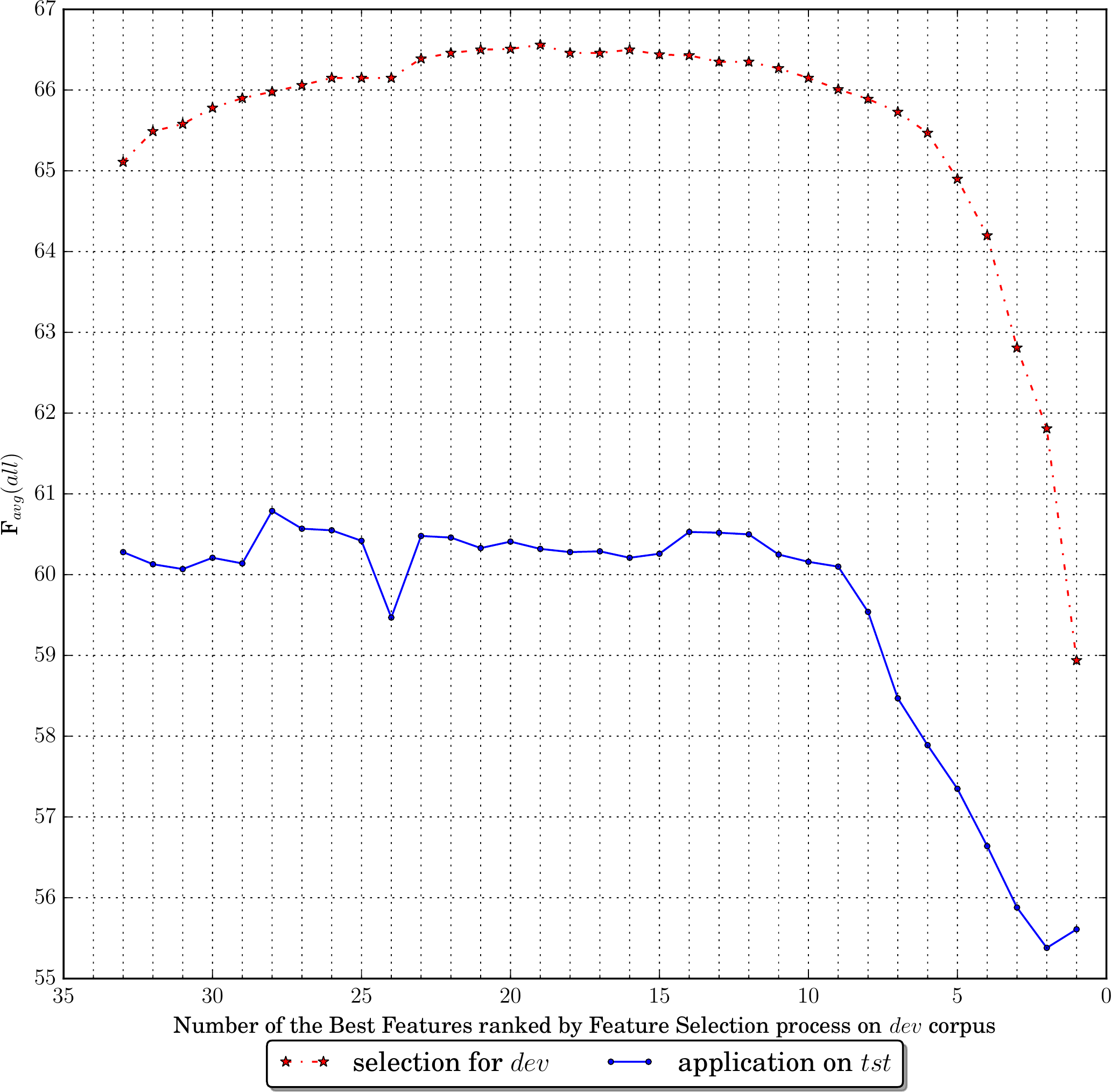}
\caption{Evolution of WCE performance for \textit{dev} (features selected) and \textit{tst} corpora when feature selection using SBS algorithm is made on \textit{dev} ($ASR1$ system).}
\label{fig:feature_selection_tgt_slt1}       
\end{figure*}

\begin{figure*}
  \includegraphics[width=0.75\textwidth]{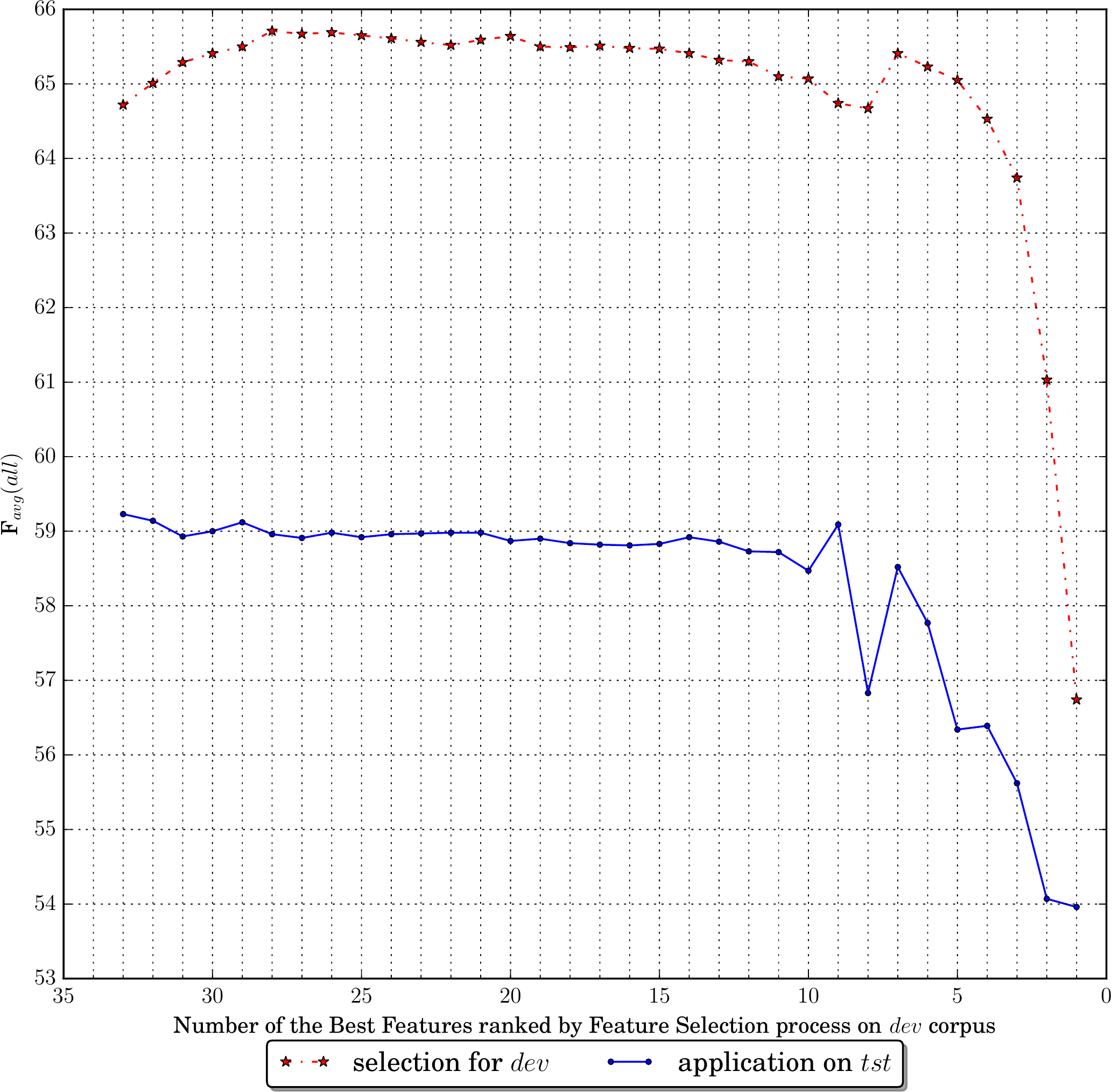}
\caption{Evolution of WCE performance for \textit{dev} (features selected) and \textit{tst} corpora when feature selection using SBS algorithm is made on \textit{dev} ($ASR2$ system).}
\label{fig:feature_selection_tgt_slt2}       
\end{figure*}

Figure \ref{fig:feature_selection_tgt_slt1} and Figure \ref{fig:feature_selection_tgt_slt2} present  the evolution of WCE performance for \textit{dev}  and \textit{tst} corpora when feature selection using SBS algorithm is made on \textit{dev}, for $ASR1$ and $ASR2$ systems, respectively. In other words, for these two figures, we apply our SBS algorithm on \textit{dev} which means that feature selection is done on \textit{dev} with classifiers trained on \textit{tst}. After that, the best feature subsets (using 33, 32, 31 until 1 feature only) are applied on \textit{tst} corpus (with classifiers trained on \textit{dev})\footnote{3 data sets would have been needed to (a) train classifiers, (b) apply feature selection, (c) evaluate WCE performance. Since we only have a \textit{dev} and a \textit{tst} set, we found this procedure acceptable}.   

On both figures, we observe that half of the features only contribute to the WCE process since best performances are observed with 10 to 15 features only. We also notice that optimal WCE performance is not necessarily obtained with the full feature set but it can be obtained with a subset of it.

\section{Conclusion}
\label{sec:illust}

\subsection{Main contributions}

In this paper, we introduced a new quality assessment task: word confidence estimation (WCE) for spoken language translation (SLT). A specific corpus, distributed to the research community\footnote{\url{https://github.com/besacier/WCE-SLT-LIG}} was built for this purpose. 
We  formalized WCE for SLT and proposed several approaches based on several types of features: machine translation (MT) based features, automatic speech recognition (ASR) based features, as well as combined or joint features using ASR and MT information. The proposition of a unique \textit{joint} classifier based on different feature types (ASR and MT features) allowed us to operate feature selection and analyze which features (from ASR or MT) are the most efficient for quality assessment in speech translation. Our conclusion is that MT features remain the most influential
while ASR feature can bring interesting complementary information. In all our experiments, we  systematically evaluated with two ASR systems that have different performance in order to analyze the behavior of our quality assessment algorithms at different levels of word error rate (WER). This allowed us to observe that WCE performance decreases as ASR system improves.
For reproducible research,
most features\footnote{MT features already available, ASR features available soon} and algorithms used in this paper are available through our toolkit called WCE-LIG. This package is made 
available on a \textit{GitHub} repository\footnote{\url{https://github.com/besacier/WCE-LIG}} under the licence GPL V3. We hope that the availability of our corpus and toolkit could lead, in a near future, to a new shared task dedicated to quality estimation for speech translation. Such a shared task could be proposed in avenues such as IWSLT (International Workshop on Spoken Language Translation) or WMT (Workshop on Machine Translation) for instance.

\subsection{SLT redecoding using WCE}

A direct application of this work is the use of WCE labels to re-decode speech translation graphs and (hopefully) improve speech translation performance. Preliminary results were already obtained and recently published by the authors of this paper  \cite{besacier-asru-2015}. 
The main idea is to carry a second speech translation pass by considering every word and its quality assessment label, as shown in \textit{equation 4}. The speech translation graph is redecoded following the following principle: words labeled as $good$ in the search graph should be ``rewarded'' by reducing their cost; on the contrary, those labeled as $bad$ should be ``penalized''. 
To illustrate this direct application of our work, we present examples of speech translation hypotheses (SLT) obtained with or without graph re-decoding  in table \ref{tab:examples} (table taken from \cite{besacier-asru-2015}).

Example 1 illustrates a first case where re-decoding  allows slightly improving the translation hypothesis.
Analysis of the labels from the confidence estimator indicates that the words \textit{a} (start of sentence) and \textit{penalty}  were labeled as $bad$ here. Thus, a better hypothesis arised from the second pass, although the transcription error could not be recovered.
In example 2, the confidence estimator labeled as $bad$ the following word sequences: \textit{it has}, \textit{speech that was} and \textit{post route}. Better translation hypothesis is found after re-decoding (correct pronoun, better quality at the end of sentence).
Finally, example 3 shows a case where, this time, the end of the first pass translation deteriorated after re-decoding. Analysis of confidence estimator output shows that the phrase \textit{to open} was (correctly) labeled as $bad$, but the re-decoding gave rise to an even worse hypothesis. The reason is that the system could not recover the named entity \textit{opel} since this word was not in the speech translation graph.

\begin{table}
\caption{Examples of French SLT hyp with and w/o graph re-decoding - table taken from \cite{besacier-asru-2015}.}
\label{tab:examples}       
\begin{tabular}{p{3.2cm}p{7.8cm}}
\hline\noalign{\smallskip}
\textit{$f_{ref_{1}}$} & une démobilisation des employés peut déboucher sur une démoralisation \bf mortifère \\
\textit{$f_{hyp_{1}}$} & une démobilisation des employés peut déboucher sur une démoralisation \bf{mort y faire} \\
\textit{$e_{hyp_{1}}$} baseline & a \textbf{demobilisation employees} can lead to a \textbf{penalty demoralisation} \\
\textit{$e_{hyp_{1}}$} with re-decoding & a \textbf{demobilisation of employees} can lead to a \textbf{demoralization death} \\
\textit{$e_{ref_{1}}$}& \textbf{demobilization of employees} can lead to a \textbf {deadly demoralization} \\
\hline\noalign{\smallskip}
\hline\noalign{\smallskip}
\textit{$f_{ref_{2}}$} & celui-ci a indiqué que l'intervention \textbf{s'}était parfaitement bien \textbf{déroulée} et que les examens post-\textbf{opératoires} étaient normaux \\
\textit{$f_{hyp_{2}}$} & celui-ci a indiqué que l' intervention \textbf{c'}était parfaitement bien \textbf{déroulés} , et que les examens post \textbf{opératoire} étaient normaux. \\
\textit{$e_{hyp_{2}}$} baseline & \textbf{it} has indicated that the speech \textbf{that was well} conducted , and that the tests were \textbf{normal post route}	 \\
\textit{$e_{hyp_{2}}$} with re-decoding & \textbf{he} indicated that the intervention \textbf{is very well done} , and that the tests \textbf{after operating were normal} \\
\textit{$e_{ref_{2}}$} & \textbf{he} indicated that the operation \textbf{went perfectly well} and the \textbf{post-operative tests were normal} \\

\hline\noalign{\smallskip}
\hline\noalign{\smallskip}
\textit{$f_{ref_{3}}$} & general motors repousse jusqu’en janvier le plan pour \textbf{opel} \\
\textit{$f_{hyp_{3}}$} & general motors repousse jusqu' en janvier le plan pour \textbf{open} \\
\textit{$e_{hyp_{3}}$} baseline & general motors postponed until january \textbf{the plan to open}	 \\
\textit{$e_{hyp_{3}}$} with re-decoding & general motors puts until january \textbf{terms to open} \\
\textit{$e_{ref_{3}}$} & general motors postponed until january \textbf{the plan for opel} \\
\noalign{\smallskip}\hline
\end{tabular}
\end{table}

\subsection{Other perspectives}

In addition to re-decode SLT graphs, our quality assessment system can be used in interactive speech translation scenarios such as news or lectures subtitling, to improve human translator productivity by giving him/her feedback on automatic transcription and translation quality. Another application would be the adaptation of our WCE system to interactive speech-to-speech translation scenarios where feedback on transcription and translation modules is needed to improve communication. On these latter subjects, it would also be nice to move from a binary ($good$ or $bad$ labels) to a 3-class decision problem (\textit{good, asr-error, mt-error}). The outcome material of this paper (corpus, toolkit) can be definitely used to address such a new problem.

\bibliographystyle{spmpsci}      
\bibliography{slt_mt_journalbib}


\end{document}